%% file: ms.tex
\documentclass{llncs}
\usepackage{makeidx}  \usepackage[dvipsnames,table,xcdraw]{xcolor}
\usepackage[pagebackref=false,breaklinks=true,colorlinks]{hyperref}
\usepackage{url}
\usepackage{graphicx}

\graphicspath{{./figures/}}

\usepackage{float}
\usepackage{amsfonts}
\usepackage{amsmath}
\usepackage{amssymb}
\usepackage{xspace}
\usepackage{bm}
\usepackage{booktabs}
\usepackage{inputenc}
\usepackage{fancyhdr}

\makeatletter
\DeclareRobustCommand\onedot{\futurelet\@let@token\@onedot}
\def\@onedot{\ifx\@let@token.\else.\null\fi\xspace}
\def\eg{\emph{e.g}\onedot} 
\def\ie{\emph{i.e}\onedot} 
 
\def\etc{\emph{etc}\onedot}

\makeatother
\def\fig#1{Figure~\ref{fig:#1}}
\def\tab#1{Table~\ref{tab:#1}}
\def\sect#1{Section~\ref{sec:#1}}

\def\bea{\begin{eqnarray}}
\def\eea{\end{eqnarray}}

\def\Eq#1{Eq.~(\ref{eq:#1})}
\def\eq#1{(\ref{eq:#1})}

\def\ex#1#2{\textrm{I\!E}_{#1}\!\left[#2\right]}                                   
\def\bx{{\bf x}}
\def\by{{\bf y}}
\def\bz{{\bf z}}
\def\bt{{\bm \theta}}
\def\bp{{\bm \phi}}

\restylefloat{table}
\fancypagestyle{firststyle}
{
   \fancyhf{}
      \fancyfoot[L]{\footnotesize Published as a conference paper at ECML-PKDD 2018}
   \renewcommand{\footrulewidth}{1pt}
   \renewcommand{\headrulewidth}{0pt}
}

\begin{document}
\title{Auxiliary Guided Autoregressive Variational Autoencoders}

\titlerunning{Auxiliary Guided Autoregressive Variational Autoencoders}  \author{Thomas Lucas\and Jakob Verbeek}
\authorrunning{Thomas LUCAS and Jakob Verbeek} \tocauthor{Thomas Lucas, Jakob Verbeek}

\institute{Universit\'{e}.\ Grenoble Alpes, Inria, CNRS, Grenoble INP, LJK\\
38000 Grenoble, France\\
\email{\texttt{\{name.surname\}@inria.fr}},\\ 
}

\maketitle              \thispagestyle{firststyle}

\newcommand{\Z}{\mathcal{Z}}
\newcommand{\red}[1]{#1}
\newcommand{\ora}[1]{\textcolor{BurntOrange}{#1}}
\newcommand{\gre}[1]{\textcolor{green}{#1}}
\newcommand{\ble}[1]{#1}
\newcommand{\mcheck}{\checkmark}
\newcommand{\nocheck}{$ $}
\newcommand{\citep}[1]{\cite{#1}}
\newcommand{\citet}[1]{\cite{#1}}

\newcommand{\effectOfKL}{
	\begin{figure}[t]		\begin{minipage}[b]{0.48\textwidth}
				\begin{center}
					{\small						$\lambda = 2$}
					\input{./figures/dec_mismatch0.tex} \\
					{\small						$\lambda = 8$}
					\input{./figures/dec_mismatch1.tex} \\
					{\small						$\lambda = 12$}
					\input{./figures/dec_mismatch2.tex}\\
				\vspace{0.1cm}
				(a)
				\end{center}
						\end{minipage}		\hspace{2mm}
		\begin{minipage}[b]{0.48\textwidth}
				\begin{center}
					{\small						$\lambda = 2$}
					\input{./figures/resp_cond_v20.tex} \\
					{\small						$\lambda = 8$}
					\input{./figures/resp_cond_v21.tex} \\
					{\small						$\lambda = 12$}
					\input{./figures/resp_cond_v22.tex}\\
				\vspace{0.1cm}
				(b)
				\end{center}
						\end{minipage}				\caption{ 
		Effect of the regularization parameter $\lambda$.
		Reconstructions (a) and samples (b) of the VAE decoder (VR and VS, respectively) and corresponding conditional samples from the pixelCNN (PS).
		}
		\label{fig:lambda_samples}
	\end{figure}
}

\newcommand{\quantitativeEval}{
	\begin{table}[t]
		\begin{center}
		{\footnotesize		\begin{tabular}{lrrrr}		\toprule 
		Model  & & BPD & $ .| z$ & $ .| x_{j<i}$ \\
		\midrule
		NICE  &  \citep{dinh15iclr}  & 4.48 & \mcheck & \nocheck \\
		Conv.\ DRAW &\citep{gregor16nips}  & $\leq$ 3.58 & \mcheck & \nocheck \\
		Real NVP  &\citep{dinh17iclr} & 3.49 & \mcheck &  \nocheck \\
		MatNet &\citep{bachman16nips}  & $\leq$ 3.24 & \mcheck & \nocheck \\
		PixelCNN &\citep{oord16icml}  & 3.14 & \nocheck & \mcheck \\
		VAE-IAF &\citep{kingma16nips} & $ \leq$ 3.11 & \mcheck &  \nocheck \\
		Gated pixelCNN & \citep{oord16nips} & 3.03 &  \nocheck & \mcheck  \\
		Pixel-RNN &\citep{oord16icml} & 3.00 & \nocheck & \mcheck \\
		Aux.\ pixelCNN &\citep{kolesnikov17icml} & 2.98 & \nocheck & \mcheck \\ 
		Lossy VAE &\citep{chen17iclr} & $\leq$ 2.95 & \mcheck & \mcheck \\
		{\bf AGAVE}, $\lambda=12$  & (this paper) &  $ \leq$ 2.92 & \mcheck & \mcheck  \\ 
		pixCNN++  &\citep{salimans17iclr} & 2.92 & \nocheck & \mcheck \\
		\bottomrule
		\end{tabular}
		}
		\end{center}
		\caption{Bits per dimension (lower is better) of models on the CIFAR10 test data.}
		\label{tab:state_of_art}
	\end{table}
}

\newcommand{\klCurves}{
	\begin{figure}[t]
		\begin{center}
			\includegraphics[width=.5\linewidth]{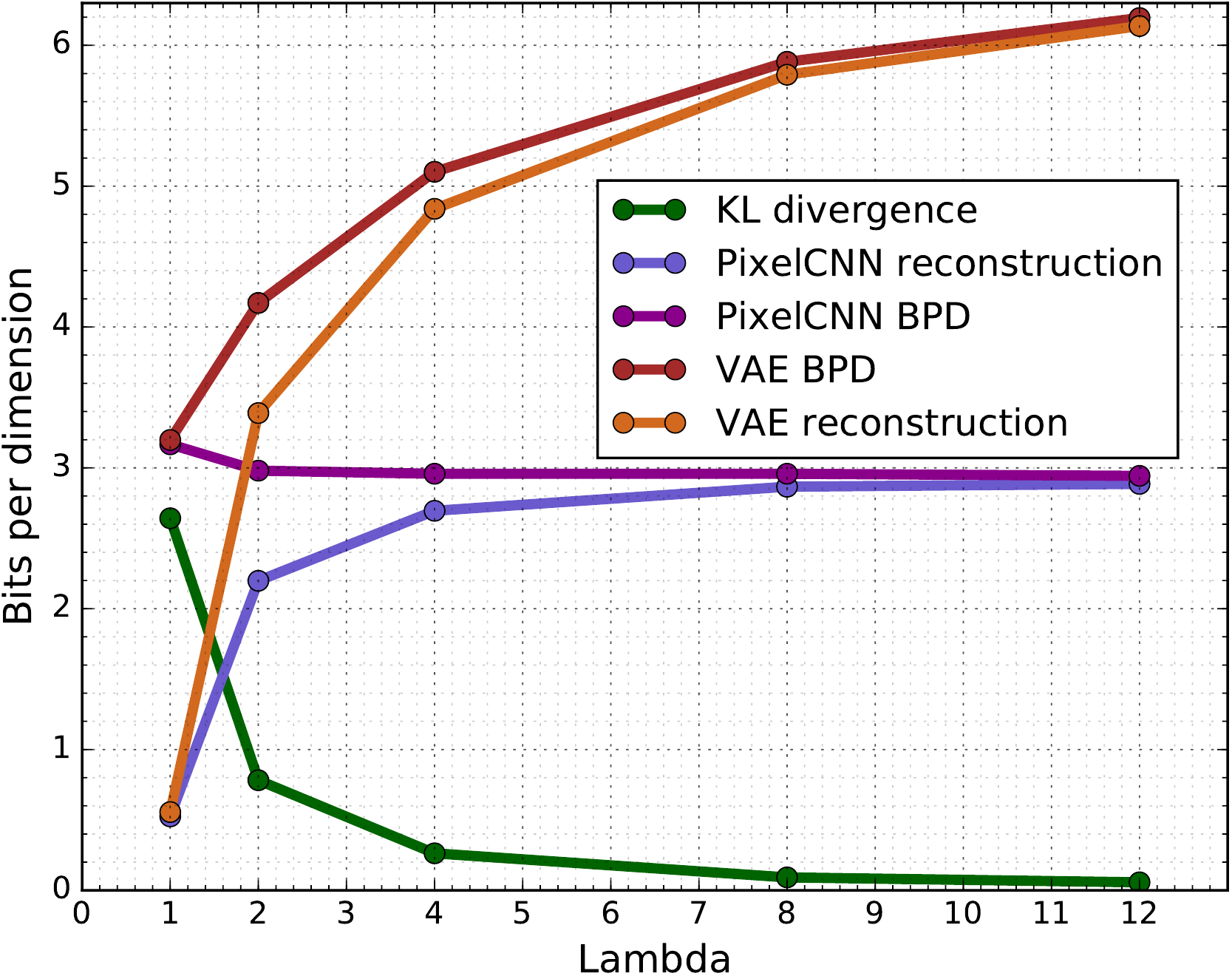}\\
		\end{center}
		\caption{
			Bits per dimension of the VAE decoder and pixelCNN decoder, as well as decomposition in KL regularization and reconstruction terms.
		}
		\label{fig:lambda_curve}
	\end{figure}
}

\newcommand{\controlFig}{
	\begin{figure}[t]
		\begin{center}
			\input{./figures/test_rebuttal_control_rec_small.tex}
		\end{center}
	       \caption{Auxiliary reconstructions obtained after dropping the auxilliary loss. (GT) denotes ground truth images unseen during training, $f(z)$ is the corresponding intermediate reconstruction, (PS) denotes pixelCNN samples, conditionned on $f(z)$.}
	       \label{fig:control_reconstructions}
	\end{figure}
}

\newcommand{\quantizeFig}{
	\begin{figure}[]
		\begin{center}
			\begin{minipage}[b]{0.48\textwidth}
				\begin{center}
					\scalebox{1.1}{
					\input{./figures/fig5a.tex}}\\
					(a)
				\end{center}
			\end{minipage}			\hfill
			\begin{minipage}[b]{0.45\textwidth}
				\centering
				\includegraphics[width=\linewidth]{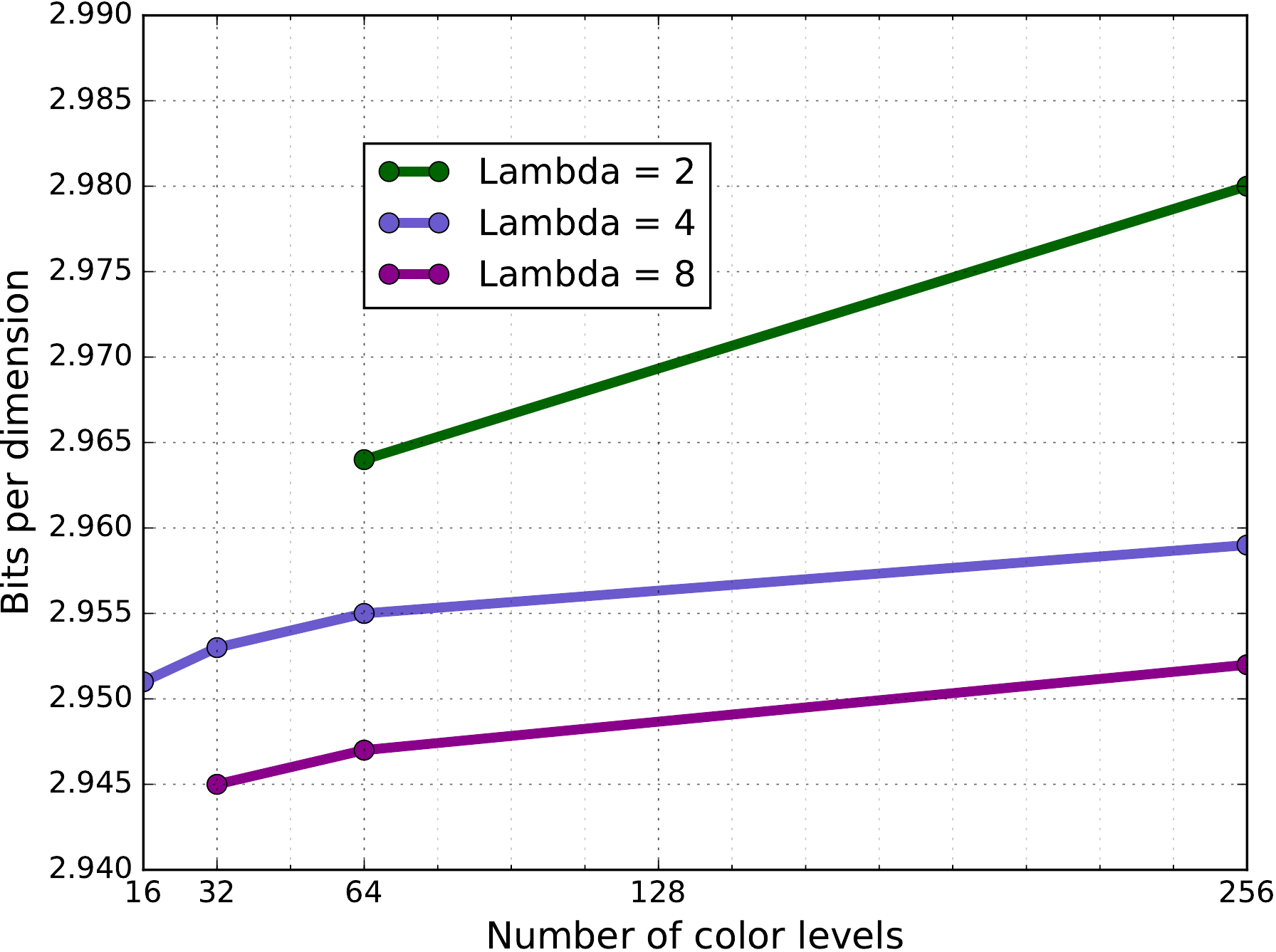}\\
				(b)
			\end{minipage}
		\end{center}
		\caption{
			Impact of the color quantization in the auxiliary image. 
			(a) Reconstructions of the VAE decoder for different quantization levels ($\lambda=8$).
			(b) BPD as a function of the quantization level.
		}
		\label{fig:bin_plots}
	\end{figure}
}

\newcommand{\scaleFig}{
	\begin{figure}[]
		\begin{center}
			\begin{minipage}[b]{0.48\textwidth}
				\begin{center}
																				\input{./figures/grey_images_latex_tile_grey.tex}\\
					(a)
				\end{center}
			\end{minipage}
			\hfill
			\begin{minipage}[b]{0.48\textwidth}
				\begin{center}
																				\input{./figures/bin_samples_pix0_small.tex}\\
					(b)
				\end{center}
			\end{minipage}
		\end{center}
		\begin{center}
			\begin{minipage}[b]{0.48\textwidth}
				\begin{center}
					\input{./figures/coarse_16_new_small.tex}\\
																				(c)
				\end{center}
			\end{minipage}
			\hfill
			\begin{minipage}[b]{0.48\textwidth}
				\begin{center}
										\input{./figures/pixout_new8x_pixVAE.tex}\\
					(d)
				\end{center}
			\end{minipage}
		\end{center}
				\caption{
			Samples from models trained with grayscale auxiliary images with 16 color levels (a), 32$\times$32 auxiliary images with 32 color levels (b), and at reduced resolutions of 16$\times$16 (c) and 8$\times$8 pixels (d) with 256 color levels.
			For each model the auxilliary representation $f(\bz)$, with $\bz$ sampled from the prior, is displayed above the corresponding conditional pixelCNN sample.
		}
		\label{fig:bin_samples}
	\end{figure}
}

\newcommand{\multiSample}{
	\begin{figure}[t]
		\begin{center}
			\input{./figures/test_rebuttal_array_small.tex}
		\caption{The column labeled $f(\bz)$ displays auxiliary representations, with $\bz$ sampled from the unit Gaussian prior $p(\bz)$, accompanied by ten samples of the conditional pixelCNN.}
				\label{fig:extra_samples}
		\end{center}

	\end{figure}
}

\newcommand{\FigSamplesIntro}{
\begin{figure}[t]
\begin{center}
	\begin{minipage}[b]{0.28\textwidth}
	\centering
		\input{./figures/vae_kl1_samplesvae_kl1.tex}\\
				(a)
	\end{minipage}
	\hfill
	\begin{minipage}[b]{0.28\textwidth}
	\centering
		\input{./figures/pixcnnpp_samplespixcnn_pp.tex}\\
				(b)
	\end{minipage}
	\hfill
	\begin{minipage}[b]{0.41\textwidth}
				\centering
				\input{./figures/pix_out_samples_figure_kl8_bin32_small.tex}\\
		(c)
	\end{minipage}
\end{center}
\caption{ Randomly selected samples from unsupervised models trained on 32$\times$32 CIFAR10 images: (a) IAF-VAE \cite{kingma16nips}, (b) pixelCNN++ \cite{salimans17iclr}, and (c) our hybrid AGAVE model.
For our model, we show the intermediate high-level representation based on latent variables (left), 
that conditions the final sample based on the pixelCNN decoder (right). 
}
\label{fig:SamplesIntro}
\end{figure}
}

\newcommand{\FigModelIntro}{
	\begin{figure}[t]
	\includegraphics[angle=270, width=\linewidth]{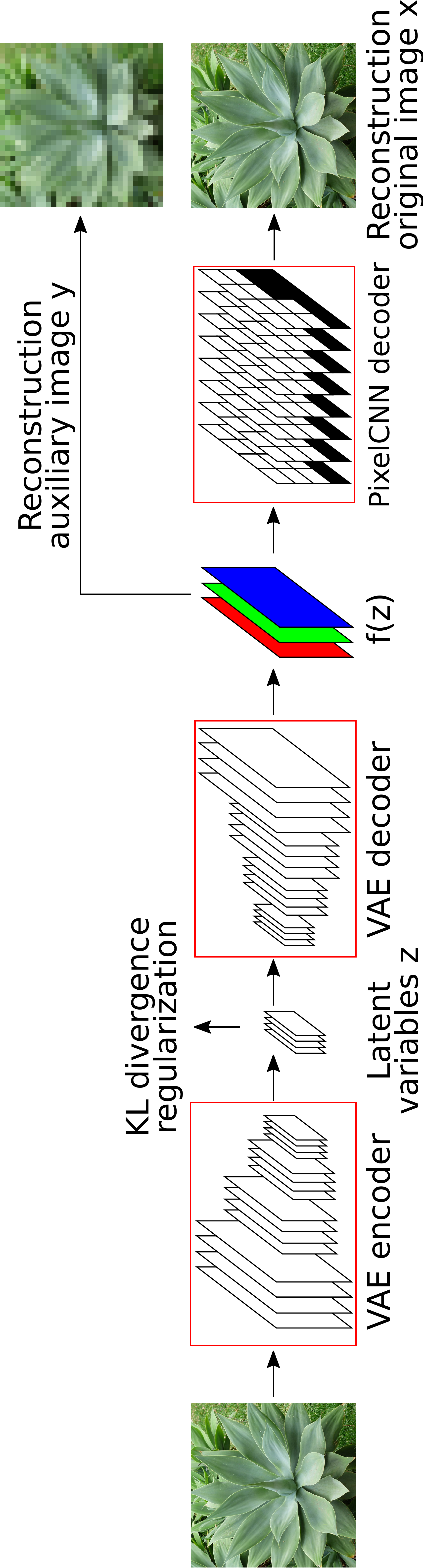}
	\caption{Schematic illustration of our auxiliary guided autoregressive variational autoencoder (AGAVE).
	The objective function has three components: KL divergence regularization, per-pixel reconstruction with the VAE decoder, and autoregressive reconstruction with the pixelCNN decoder.
	}
	\label{fig:ModelIntro}
	\end{figure}
}

\newcommand{\FigLatent}{
	\begin{figure}[t]
				\begin{center}
						\input{./figures/pixVAE_images_latent_interpolation_latex_tile_both.tex}
		\caption{The first and last columns contain auxilliary reconstructions, images in between are obtained from interpolation of the corresponding latent variables. Odd rows contain auxilliary reconstructions, and even rows contain outputs of the full model.}
		\label{fig:figLatent}
		\end{center}
	\end{figure}
}

\newcommand{\FigGrey}{
	\begin{figure}[t]
				\begin{center}
						\input{./figures/pixvae_images_grey_images_latex_tile_grey.tex}
		\caption{The first and last columns contain auxilliary reconstructions, images in between are obtained from interpolation of the corresponding latent variables. Odd rows contain auxilliary reconstructions, and even rows contain outputs of the full model.}
		\label{fig:figGrey}
		\end{center}
	\end{figure}
}

\begin{abstract}
\ble{
Generative modeling of high-dimensional data is a key problem in machine learning. Successful approaches include latent variable models and autoregressive models. The complementary strengths of these approaches, to model global and local image statistics respectively, suggest hybrid models that encode global image structure into latent variables while autoregressively modeling low level detail. 
Previous approaches to such hybrid models restrict the capacity of the autoregressive decoder to prevent degenerate models that ignore the latent variables and only rely on autoregressive modeling. Our contribution is a training procedure relying on an auxiliary loss function that controls which information is captured by the latent variables and what is left to the autoregressive decoder. Our approach can leverage arbitrarily powerful autoregressive decoders, achieves state-of-the art quantitative performance among models with latent variables, and generates qualitatively convincing samples.
}
\end{abstract}

\pagestyle{fancy}
\renewcommand{\footrulewidth}{0pt}
\renewcommand{\headrulewidth}{0pt}

\section{Introduction}
\FigModelIntro{}

Unsupervised modeling of complex distributions with unknown structure is a landmark challenge in machine learning. 
The problem is often studied in the context of learning generative models of 
the complex high-dimensional distributions of natural image collections. 
Latent variable approaches can learn disentangled and concise representations of the data \citep{bengio13pami}, which are useful for compression \citep{gregor16nips} and semi-supervised learning \citep{kingma14nips,rasmus15nips}. 
When conditioned on prior information, generative models can be used for a variety of tasks, such as attribute or class-conditional image generation, text and pose-based image generation, image colorization, \etc \citep{yan16eccv,oord16nips,reed17icml,desphande17cvpr}. 
Recently significant advances in generative (image) modeling have been made 
along several lines, including adversarial networks \citep{goodfellow14nips,arjovsky17icml}, variational autoencoders \citep{kingma14iclr,rezende14icml}, autoregressive models \citep{oord16icml,reed17icml}, and non-volume preserving variable transformations \citep{dinh17iclr}.

In our work we seek to combine the merits of two of these lines of work.
Variational autoencoders (VAEs) \citep{kingma14iclr,rezende14icml} 
can learn latent variable representations that abstract away from low-level details, 
but model  pixels as conditionally independent given the latent  variables.
This renders the generative model computationally efficient, 
but the lack of low-level structure modeling leads to overly smooth and blurry samples.
Autoregressive  models, such as pixelCNNs \citep{oord16icml}, 
on the other hand, estimate complex translation invariant conditional distributions among pixels.
They are effective to model low-level image statistics, and yield state-of-the-art likelihoods on test data \citep{salimans17iclr}.
This is in line with the observations of \citet{kolesnikov17icml} that low-level image details account for a large part of the likelihood.
These autoregressive models, however, do not learn a latent variable  representations to support, \eg, semi-supervised learning. 
The complementary strengths of VAEs and pixelCNNs, modeling global and local image statistics respectively, suggest hybrid approaches combining the strengths of both. Prior work on such hybrid models needed to limit the capacity of the autoregressive decoder to  prevent degenerate models that completely ignore the latent variables and rely on autoregressive modeling only \citep{gulrajani17iclr,chen17iclr}.
In this paper we describe Auxiliary Guided Autoregressive Variational autoEncoders (AGAVE), an approach to train such hybrid models using an auxiliary loss function that controls which information is captured by the latent variables and what is left to the AR decoder. \ble{That removes the need to limit the capacity of the latter. See \fig{ModelIntro} for a schematic illustration of our approach.} 

Using high-capacity VAE and autoregressive components allows our models to obtain quantitative results on held-out data that are on par with the state of the art \ble{in general, and set a new state of the art among models with latent variables}. Our models generate samples with  both global coherence and low-level details.  See \fig{SamplesIntro} for representative samples of VAE and pixelCNN models. \\

\section{Related work}
\label{sec:related}

\begin{figure}[t]
\begin{center}
	\begin{minipage}[b]{0.28\textwidth}
	\centering
		\input{./figures/vae_kl1_samplesvae_kl1.tex}\\
				(a)
	\end{minipage}
	\hfill
	\begin{minipage}[b]{0.28\textwidth}
	\centering
		\input{./figures/pixcnnpp_samplespixcnn_pp.tex}\\
				(b)
	\end{minipage}
	\hfill
	\begin{minipage}[b]{0.41\textwidth}
				\centering
				\input{./figures/pix_out_samples_figure_kl8_bin32_small.tex}\\
		(c)
	\end{minipage}
\end{center}
\caption{ Randomly selected samples from unsupervised models trained on 32$\times$32 CIFAR10 images: (a) IAF-VAE \cite{kingma16nips}, (b) pixelCNN++ \cite{salimans17iclr}, and (c) our hybrid AGAVE model.
For our model, we show the intermediate high-level representation based on latent variables (left), 
that conditions the final sample based on the pixelCNN decoder (right). 
}
\label{fig:SamplesIntro}
\end{figure}
{}

Generative image modeling has recently taken significant strides forward, 
leveraging deep neural networks to learn complex density models using a variety of approaches.
These include the variational autoencoders and autoregressive models that form the basis of our work, but also generative adversarial networks (GANs) \citep{goodfellow14nips,arjovsky17icml} and variable transformation with invertible functions \citep{dinh17iclr}.
While GANs produce visually appealing samples, they suffer from mode dropping and their likelihood-free nature prevents measuring how well they model held-out test data. 
In particular, GANs can only generate samples on a non-linear manifold in the data space  with dimension equal to the number of latent variables. 
In contrast, probabilistic models such as VAEs and autoregressive models generalize to the entire data space, and likelihoods of held-out data can be used for  compression, and to quantitatively compare different models.
The non-volume preserving (NVP) transformation approach of \citet{dinh17iclr} chains together invertible transformations to map a basic  (\eg unit Gaussian) prior on the latent space to a complex distribution on the data space. 
This method offers tractable likelihood evaluation and exact inference, but obtains likelihoods on held-out data below the values reported using state-of-the-art VAE and autoregressive models.
Moreover, it is restricted to use latent representations with the same dimensionality as the input data, and is thus difficult to scale to model high-resolution images.

Autoregressive density estimation models, such as pixelCNNs \citep{oord16icml}, 
admit tractable likelihood evaluation, while for variational autoencoders 
\citep{kingma14iclr,rezende14icml} accurate approximations can be obtained using importance sampling \citep{burda16iclr}.
Naively combining powerful pixelCNN decoders in a VAE framework results in a degenerate model which ignores the VAE latent variable structure, 
as explained through the lens of bits-back coding by \citet{chen17iclr}.
To address this issue, the capacity of the the autoregressive component can be restricted. This can, for example, be achieved  by reducing its depth and/or field of view, or by giving the pixelCNN only access to grayscale values, \ie modeling $p(x_i|\bx_{<i},\bz)= p(x_i|\text{gray}(\bx_{<i}),\bz)$  \citep{chen17iclr,gulrajani17iclr}. 
This forces the model to leverage the latent variables $\bz$ to model part of the dependencies among the pixels.
This approach, however, has  two drawbacks. 
(i) Curbing the capacity of the model is undesirable in unsupervised settings where training data is abundant and overfitting unlikely, \ble{and is only a partial solution to the problem.}
(ii) Balancing what is modeled by the VAE and the pixelCNN by means of architectural design choices requires careful hand-design and tuning of the architectures. \ble{This is a tedious process, and a more reliable principle is desirable.}
To overcome these drawbacks, we propose to instead control what is modeled by the VAE and pixelCNN with an auxiliary loss on the VAE decoder output before it is used to condition the autoregressive  decoder. 
This allows us to ``plug in'' powerful high-capacity VAE and pixelCNN architectures, and balance what is  modeled by each component by means of the auxiliary loss.

In a similar vein, \citet{kolesnikov17icml}  force pixelCNN models to capture more high-level image aspects using  an auxiliary representation $\by$ of the original image $\bx$, \eg a low-resolution version of the original. They learn a pixelCNN for $\by$, and a conditional pixelCNN to predict $\bx$ from $\by$, possibly using several intermediate representations.  
This  approach forces modeling of more high-level aspects in the intermediate representations, and yields visually more compelling samples. 
\cite{reed17icml} similarly learn a series of conditional autoregressive models to upsample coarser intermediate latent images. By introducing partial conditional independencies in the model they scale the model to efficiently sample high-resolution images of up to 512$\times$512  pixels. 
\citet{gregor16nips} use a recurrent VAE model to produces a sequence of RGB images with increasing detail derived from latent variables associated with each iteration. 
Like our work, all these models work with intermediate representations in RGB space to learn accurate generative image models.

\section{Auxiliary guided autoregressive variational autoencoders}

We give a brief overview of variational autoencoders and their limitations in \sect{vae}, 
before we present our approach to learning variational autoencoders with autoregressive decoders in \sect{agave}.

\subsection{Variational autoencoders}
\label{sec:vae}

Variational autoencoders \citep{kingma14iclr,rezende14icml} learn deep generative latent variable models using two neural networks. 
The ``decoder'' network implements a conditional distribution $p_{\theta}(\bx|\bz)$ over observations $\bx$ given a latent variable $\bz$, with parameters $\bt$.
Together with a basic prior on the latent variable $\bz$, \eg a unit Gaussian, 
the generative model on $\bx$ is obtained by marginalizing out the latent variable:
\bea
p_\bt(\bx) = \int p(\bz)p_\bt(\bx|\bz)\; d\bz.
\eea
The marginal likelihood can, however, not be optimized directly since the non-linear dependencies in  $p_{\bt}(\bx|\bz)$ render the integral intractable. To overcome this problem, an ``encoder'' network is used to compute an approximate posterior distribution $q_\bp(\bz|\bx)$, with parameters $\bp$. 
The approximate posterior is used to define a variational bound on the data log-likelihood, by subtracting the Kullback-Leibler divergence between the true and approximate posterior:
\bea
\ln p_{\bt}(\bx) \geq \mathcal{L}(\bt,\bp;\bx)& = & 		\ln(p_{\bt}(\bx)) - D_\text{KL}(q_{\bp}(\bz|\bx) || p_{\bt}(\bz|\bx))\label{eq:elbo}\\ 
 & = & \underbrace{\ex{q_{\bp}}{\ln(p_{\bt}(\bx|\bz)}}_{\text{Reconstruction}} - \underbrace{D_\text{KL}(q_{\bp}(\bz|\bx) || p(\bz))}_{\text{Regularization}}.\label{eq:recreg}
\eea
The decomposition in \eq{recreg} interprets the bound as the sum of a reconstruction term and a regularization term. The first aims to maximize the expected data log-likelihood $p_\bt(\bx|\bz)$ given the posterior estimate $q_\bp(\bz|\bx)$.
The second term prevents $q_\bp(\bz|\bx)$ from collapsing to a single point, which would be optimal for the first term.

Variational autoencoders typically model the dimensions of $\bx$ as conditionally independent, \begin{equation}
p_\bt(\bx|\bz)=\prod_{i=1}^D p_\bt(x_i|\bz),
\end{equation}
for instance using a factored Gaussian or Bernoulli model, 
see \eg  \cite{kingma14iclr,kingma16nips,yan16eccv}. 
The conditional independence assumption makes sampling from the VAE efficient:
since the decoder network is evaluated only once for a sample $\bz\sim p(\bz)$ to compute  
all the conditional distributions $p_\bt(x_i|\bz)$, the $x_i$ can then be sampled in parallel.

A result of relying on the latent variables to account for all pixel dependencies,  however, is that all low-level variability must also be modeled by the latent variables.
Consider, for instance, a picture of a dog, and variants of that image shifted by one or a few pixels, or in a slightly different pose, with a slightly lighter background,  or with less saturated colors, \etc. 
If these factors of variability are modeled using latent variables, then these low-level aspects are confounded with latent variables relating to the high-level image content.
If the corresponding image variability is not modeled using latent variables, it will be modeled as independent pixel noise. 
In the latter case, using the mean of $p_\bt(\bx|\bz)$ as the synthetic image for a given $\bz$  results in blurry samples, since the mean is averaged over the low-level variants of the image.  
Sampling from $p_\bt(\bx|\bz)$ to obtain synthetic images, on the other hand,  results in images with unrealistic independent pixel noise.

\subsection{Autoregressive decoders in variational autoencoders}
\label{sec:agave}

Autoregressive density models, see \eg \citep{larochelle11aistats,germain15icml},
rely on the basic factorization  of multi-variate distributions,
\bea
	p_{\bt}(\bx) = \prod_{i=1}^D p_{\bt}(x_i | \bx_{<i}) 
\eea
with $\bx_{<i} = x_1,\dots, x_{i-1}$, and model the conditional distributions using a (deep)  neural network. 
For image data, PixelCNNs \citep{oord16icml,oord16nips} use a scanline  pixel ordering, and model the conditional distributions using a convolution neural network.  
The convolutional filters are masked so as to ensure that the receptive fields only extend to pixels $\bx_{<i}$ when computing the conditional distribution of $x_i$.

PixelCNNs can be used as a decoder in a VAE by conditioning on the latent variable $\bz$ in addition to the preceding pixels, 
leading to a variational bound with a modified reconstruction term:
\bea
	\mathcal{L}(\bt,\bp;\bx) = \ex{q_{\bp}}{\sum_{i=1}^D \ln p_{\bt}(x_i|\bx_{<i},\bz)} - D_\text{KL}(q_{\bp}(\bz|\bx) || p(\bz)).
	\label{eq:elboAR}
\eea
The regularization term can be interpreted as a ``cost'' of using the latent variables. 
To effectively use the latent variables, the approximate posterior $q_\bp(\bz|\bx)$ must differ from the prior $p(\bz)$,  which increases the KL divergence.

\citet{chen17iclr} showed that for loss \eqref{eq:elboAR} and a decoder with enough capacity, it is optimal to encode no information about $x$ in $z$ by setting $q(z|x) = p(z)$.
To ensure meaningful latent representation learning \citet{chen17iclr} and \citet{gulrajani17iclr} restrict the capacity of the pixelCNN decoder. 
In our approach, in contrast, it is always optimal for the autoregressive decoder, regardless of its capacity, to exploit the information on $\bx$ carried by $z$. We rely on two decoders in parallel: the first one reconstructs an auxiliary image $\by$ from an intermediate representation $f_\bt(\bz)$ in a non-autoregressive manner.
The auxiliary image can be either simply taken to be the original image ($\by=\bx$), 
or a compressed version of it, \eg with lower resolution or with a coarser color quantization. The second decoder is a conditional autoregressive model that predicts $\bx$ conditioned on$f_\bt(\bz)$.  Modeling $\by$ in a non-autoregressive manner ensures a meaningful representation $\bz$ and renders $\bx$ and $\bz$ dependent, inducing a certain non-zero KL ``cost'' in \eqref{eq:elboAR}. The uncertainty on $\bx$ is thus reduced when conditioning on $\bz$, and there is no longer an advantage in ignoring the latent variable for the autoregressive decoder. \ble{We provide a more detailed explanation of why our auxiliary loss ensures a meaningful use of latent variables in powerful decoders in \ref{sec:bits_back}.}
To train the model we combine both decoders in a single objective function with a shared encoder network:
\begin{align}
	\mathcal{L}(\bt,\bp;\bx,\by) = \underbrace{\ex{q_{\bp}}{\sum_{i=1}^D \ln p_{\bt}(x_i|\bx_{<i},\bz)}}_{\text{Primary Reconstruction}} &+& 
\underbrace{\ex{q_{\bp}}{\sum_{j=1}^E \ln p_{\bt}(y_j|\bz)}}_{\text {Auxiliary Reconstruction}} \nonumber \\ 
&-& 
\underbrace{\lambda\; D_\text{KL}\left(q_{\bp}(\bz|\bx) || p(\bz)\right)}_{\text{Regularization}}.
\label{eq:agave}
\end{align}

Treating $\bx$ and $\by$ as two variables that are conditionally independent given a shared underlying latent vairable $\bz$ leads to $\lambda=1$.
Summing the lower bounds in \Eq{recreg} and \Eq{elboAR} of the marginal log-likelihoods of $\by$ and $\bx$, and sharing the encoder network, leads to $\lambda=2$. 
Larger values of $\lambda$ result in valid but less tight lower bounds of the log-likelihoods. Encouraging the variational posterior to be closer to the prior, 
this leads to less informative latent variable representations.

Sharing the encoder across the two decoders is the key of our approach. 
The factored auxiliary VAE decoder can only model pixel dependencies by means of the latent variables,
which ensures that a meaningful representation is learned.
Now, given that the VAE encoder output is informative on the image content, 
there is no incentive for the  autoregressive decoder to ignore the intermediate representation $f(\bz)$ on which it is conditioned. 
The choice of the regularization parameter $\lambda$ and auxiliary image $\by$ provide two levers to  control  \emph{how much} and \emph{what type} of information should be encoded in the latent variables.

\subsection{It is optimal for the autoregressive decoder to use $\bz$}

\label{sec:bits_back}
\ble{Combining a VAE with a flexible decoder (for instance an autoregressive one) leads to the latent code being ignored.
This problem could be attributed to optimization challenges: at the start of training $q(\bz|\bx)$ carries little information about $\bx$, 
the KL term pushes the model to set it to the prior to avoid any penalty, and training never recovers from falling into that local minimum. \cite{chen17iclr} have proposed extensive explanations showing that the problem goes deeper: if a sufficiently expressive decoder is used, ignoring the latents actually is the optimal behavior. The gist of the argument is based on bits-back coding as follows:
given an encoder $q(\bz|\bx)$, a decoder $p(\bx|\bz)$ and a prior $p(\bz)$,  $\bz \sim q(\bz|\bx)$ can be encoded in a lossless manner using $p(\bz)$, and $\bx$ can be encoded, also losslessly, using $p(\bx|\bz)$. Once the receiver has decoded $\bx$,
$q(\bz|\bx)$ becomes available and a secondary message can be decoded from it. This yields and average code length of: 
$$ 
C_{BitsBack} = \mathbb{E}_{\bx \sim D, \bz \sim q(.|\bx)}[\log(q(\bz|\bx)) -\log(p(\bz)) -\log(p(\bx|\bz))]. 
$$ 
$C_{BitsBack}$ corresponds to the standard VAE objective. A lower-bound on the expected code length for the data being encoded is given by the Shannon entropy: $\mathcal{H}(D) = \mathbb{E}_{\bx \sim D}[- \log p_{D}(\bx)]$, which yields: 
\begin{eqnarray*}
C_{BitsBack} &=& \mathbb{E}_{\bx \sim D}[-\log(p(\bx)) + D_{KL}(q(\bz|\bx)||p(\bz|\bx))] \\
	     &\geq& \mathcal{H}(D) + \mathbb{E}_{\bx \sim D}[D_{KL}(q(\bz|\bx)||p(\bz|\bx))].
\end{eqnarray*}
}

If $p(.|\bx_{j<i})$ is expressive enough, or if $q(.|\bx)$ is poor enough, the following inequality can be verified: 
$$
\mathcal{H}(D) \leq \mathbb{E}_{\bx \sim D}[- \log p(\bx|\bx_{j<i})] < \mathcal{H}(D) + \mathbb{E}_{\bx \sim D}[D_{KL}(q(\bz|\bx)||p(\bz|\bx))]
$$
\ble{This is always true in the limit of infinitely expressive autoregressive decoders.} In that case, any use of the latents that $p$ might decrease performance. The optimal behavior is to set $q(\bz|\bx) = p(\bz)$ to avoid the extra KL cost. Then $\bz$ becomes independent from $\bx$ and no information about $\bx$ is encoded in $\bz$. 
Therefore, given an encoder, the latent variables will only be used if the capacity of the autoregressive decoder is sufficiently restricted. This is the approach taken by \cite{chen17iclr} and \cite{gulrajani17iclr}. This approach works: it has obtained competitive quantitative and qualitative performance. However, it is not satisfactory in the sense that autoregressive models cannot be used to the full extent of their potential, while learning a meaningful latent variable representation.\\

In our setting, both $(Y,X)$ have to be sent to and decoded by the receiver.
Let us denote $C_{VAE}$ the expected code length required to send the auxiliary message, $\by$.  Once $\by$ has been sent, sending $\bx$ costs: $\mathbb{E}_{\bz \sim q(\bz | \bx)} [- \sum_i \log(p(x_i|\bz,\bx_{j<i}))] $, and we have:  

\begin{eqnarray}
	C_{VAE} &=& \mathbb{E}_{\bx \sim D, \bz \sim q(.|\bx)}[\log(q(\bz|\bx)) -\log(p(\bz)) -\log(p(\by|\bz))] \\
	C_{AGAVE} &=& C_{VAE}  + \mathbb{E}_{\bz \sim q(\bz | \bx)} [-\sum_i \log(p(x_i|\bz, \bx_{j<i}))]. 
\end{eqnarray}

Using the fact that the Shannon entropy is the optimal expected code length for transmitting $X|Z$, we obtain $C_{AGAVE}  \ge  C_{VAE}  + \mathcal{H}(X|Z)$ \\
The entropy of a random variable decreases when it is conditioned on another, i.e.\ $\mathcal{H}(X | Z) \le \mathcal{H}(X)$.
\ble{Therefore, the theoretical lower-bound on the expected code length in our setup is always better when the autoregressive component takes $Z$ into account, no matter its expressivity.
In the limit case of an infinitely expressive autoregressive decoder, denoted by $*$, the lower bound is attained and $C_{AGAVE}^* = C_{VAE}  + \mathcal{H}(X | Z) \le C_{VAE}  + \mathcal{H}(X).$}
In non-degenerate cases, the VAE is optimized to encode information about $X$ into a meaningful $Z$, with potentially near perfect reconstructions, and there exists $\epsilon > 0$ such that $\mathcal{H}(X | Z) < \mathcal{H}(X) - \epsilon$, making the lower bound stricly better by a possibly big margin.

This analysis shows that in our setup it is theoretically always better for the autoregressive model to make use of the latent and auxiliary representation it is conditioned on. That is true no matter how expressive the model is. It also shows that in theory our model should learn meaningful latent structure.

\section{Experimental evaluation}
\label{sec:exp}

\quantitativeEval{}

In this section we describe our experimental setup, 
and present results on CIFAR10.

\subsection{Dataset and implementation}

The CIFAR10 dataset  \citep{krizhevsky09msc} contains 6,000 images of 32$\times$32 pixels for each of the 10 object categories 
\emph{airplane, automobile, bird, cat, deer, dog, frog, horse, ship, truck}.
The images are split into 50,000 training images and 10,000 test images.
We train all our models in a completely unsupervised manner, ignoring the class information.

We implemented our model based on existing architectures.
In particular we use the VAE architecture of \citet{kingma16nips}, 
and use logistic distributions over the RGB color values.
We let the intermediate representation $f(\bz)$ output by the VAE decoder be the per-pixel and per-channel mean values of the logistics, and learn per-channel scale parameters that are used across all pixels.
The cumulative density function (CDF), given by the sigmoid function, is used to compute  probabilities across the 256 discrete color levels, 
or fewer if a lower quantization level is chosen in $\by$.
Using RGB values $y_i\in [0,255]$, we let $b$ denote the number of discrete color levels and define $c=256/b$. 
The probabilities over the $b$ discrete color levels are computed from the logistic mean and variance $\mu_i$ and $s_i$ as
\begin{equation}
p(y_i| \mu_i,s_i) = \sigma\left(c+c\lfloor y_i/c\rfloor |\mu_i,s_i\right) - \sigma\left(c\lfloor y_i/c \rfloor |\mu_i,s_i\right).
\label{eq:sigmoid}
\end{equation}

For the pixelCNN we use the architecture of \citet{salimans17iclr}, 
and modify it to be conditioned on the VAE decoder output $f(\bz)$, or possibly an upsampled version if $\by$ has a lower resolution than $\bx$. 
In particular, we apply standard non-masked convolutional layers to the VAE output, as many as there are pixelCNN layers. We allow each layer of the pixel-CNN to take additional input using non-masked convolutions from the feature stream based on the VAE output.
This ensures that the conditional pixelCNN remains autoregressive.

To speed up training, we  independently pretrain the VAE and pixelCNN in parallel, and then continue training the full model with both decoders.
We use the Adamax optimizer \citep{kingma15iclr} with a learning rate of $0.002$ without learning rate decay.
We will release our TensorFlow-based code to replicate our experiments upon publication.

	\begin{figure}[t]		\begin{minipage}[b]{0.48\textwidth}
				\begin{center}
					{\small						$\lambda = 2$}
					\input{./figures/dec_mismatch0.tex} \\
					{\small						$\lambda = 8$}
					\input{./figures/dec_mismatch1.tex} \\
					{\small						$\lambda = 12$}
					\input{./figures/dec_mismatch2.tex}\\
				\vspace{0.1cm}
				(a)
				\end{center}
						\end{minipage}		\hspace{2mm}
		\begin{minipage}[b]{0.48\textwidth}
				\begin{center}
					{\small						$\lambda = 2$}
					\input{./figures/resp_cond_v20.tex} \\
					{\small						$\lambda = 8$}
					\input{./figures/resp_cond_v21.tex} \\
					{\small						$\lambda = 12$}
					\input{./figures/resp_cond_v22.tex}\\
				\vspace{0.1cm}
				(b)
				\end{center}
						\end{minipage}				\caption{ 
		Effect of the regularization parameter $\lambda$.
		Reconstructions (a) and samples (b) of the VAE decoder (VR and VS, respectively) and corresponding conditional samples from the pixelCNN (PS).
		}
		\label{fig:lambda_samples}
	\end{figure}
{}
\subsection{Quantitative performance evaluation.}

Following previous work, we evaluate models on the test images using the bits-per-dimension (BPD) metric: the negative log-likelihood divided by the number of pixels values (3$\times$32$\times$32). It can be interpreted as the average number of bits per RGB value in a lossless compression scheme derived from the model. 

The comparison in \tab{state_of_art} shows that our model performs on par with the state-of-the-art results of the pixelCNN++ model \citep{salimans17iclr}.
Here we used the importance sampling-based bound of \citet{burda16iclr} with 150 samples to compute the BPD metric for our model.\footnote{The graphs in \fig{lambda_curve} and \fig{bin_plots} are based on  the bound in \Eq{agave} to reduce the computational effort.}
We refer to \fig{SamplesIntro} for qualitative comparison of samples from  our model and pixelCNN++, the latter generated using the publicly available code.

\subsection{Effect of KL regularization strength.}

\klCurves{}

In \fig{lambda_samples} we show reconstructions of test images and samples generated by the VAE decoder, together with their corresponding conditional pixelCNN samples for different values of $\lambda$. 
As expected, the VAE reconstructions become less accurate for larger values of $\lambda$, mainly by lacking details while preserving the global shape of the input. 
At the same time, the samples become more appealing for larger $\lambda$, suppressing the unrealistic high-frequency detail in the VAE samples obtained at lower values of $\lambda$. 
Note that the VAE samples and reconstructions become more similar as $\lambda$ increases, 
which makes the input to the pixelCNN during training and sampling more consistent.

For both reconstructions and samples, the pixelCNN clearly takes into account the output of the VAE decoder, demonstrating the effectiveness of our auxiliary loss to condition high-capacity pixelCNN decoders on latent variable representations.
Samples from the pixelCNN faithfully reproduce the global structure of the VAE output, leading to more realistic samples, in particular for higher  values of $\lambda$. 

For $\lambda=2$ the VAE reconstructions are near perfect during training, and the pixelCNN decoder does not significantly modify the appearance of the VAE output. 
For larger values of $\lambda$, the pixelCNN clearly adds significant detail to the VAE outputs.

\fig{lambda_curve} traces the BPD metrics of both the VAE and pixelCNN decoder as a function of $\lambda$. We also show the decomposition in regularization and reconstruction terms.
By increasing $\lambda$, the KL divergence can be pushed closer to zero. 
As the KL divergence term drops, the reconstruction term for the VAE rapidly increases and the VAE model obtains worse BPD values, stemming from the inability of the VAE to model pixel dependencies other than via the latent variables. 
The reconstruction term of the pixelCNN decoder also increases with $\lambda$, as the amount of information it receives drops. However, in terms of BPD which sums KL divergence and pixelCNN reconstruction, a substantial gain of 0.2 is observed increasing $\lambda$ from 1 to 2, after which smaller but consistent gains are observed.

\begin{subsection}{\red{Role of the auxilliary representation}}

\paragraph{\red{The auxilliary variables are taken into account:}}
\label{sec:more_samples}

\ble{\ref{sec:bits_back} shows that in theory it is always optimal for the autoregressive decoder to take the latent variables into account. \fig{extra_samples} demonstrates this empirically by displaying auxiliary representations $f(\bz)$ with $z$ sampled from the prior $f(z)$ as well as nine different samples from the autoregressive decoder conditioned on $f(\bz)$. This qualitatively shows that the low level detail added by the pixelCNN, which is crucial for log-likelihood performance, always respects the global structure of the image being conditioned on. The VAE decoder is trained with $\lambda = 8$ and weights very little in terms of KL divergence. Yet it controls the global structure of the samples, which shows that our setup can be used to get the best of both worlds.  
\ble{\fig{figLatent} demonstrates that the latent variables $z$ of the encoder have learned meaningfull structure with latent variable interpolations. Samples are obtained by encoding ground truth images, then interpolating the latent variables obtained, decoding them with the decoder of the $VAE$ and adding low level detail with the pixelCNN.}}

	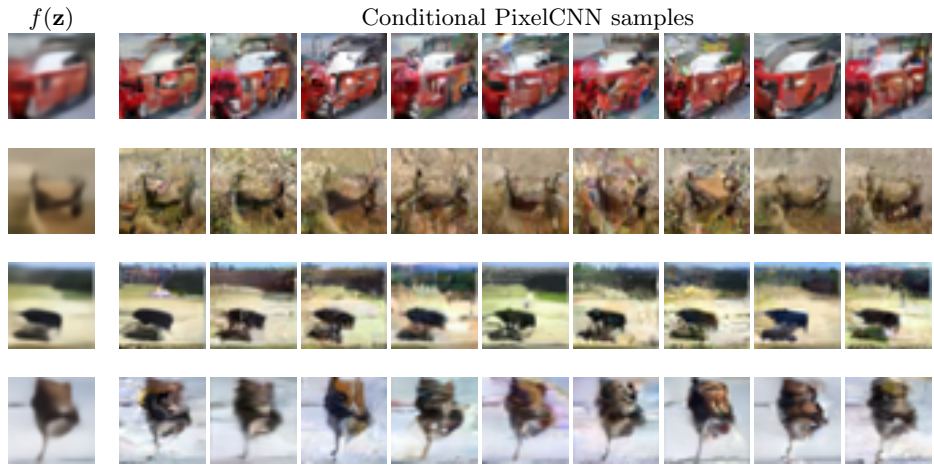
\begin{figure}[t]
		\begin{center}
			\input{./figures/test_rebuttal_array_small.tex}
		\caption{The column labeled $f(\bz)$ displays auxiliary representations, with $\bz$ sampled from the unit Gaussian prior $p(\bz)$, accompanied by ten samples of the conditional pixelCNN.}
				\label{fig:extra_samples}
		\end{center}

	\end{figure}
{}

	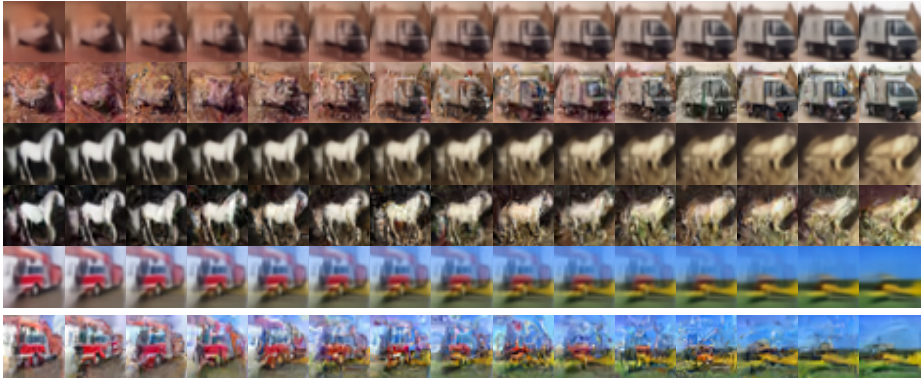
\begin{figure}[t]
				\begin{center}
						\input{./figures/pixVAE_images_latent_interpolation_latex_tile_both.tex}
		\caption{The first and last columns contain auxilliary reconstructions, images in between are obtained from interpolation of the corresponding latent variables. Odd rows contain auxilliary reconstructions, and even rows contain outputs of the full model.}
		\label{fig:figLatent}
		\end{center}
	\end{figure}
{}

\paragraph{\red{The auxilliary loss is necessary:}}

	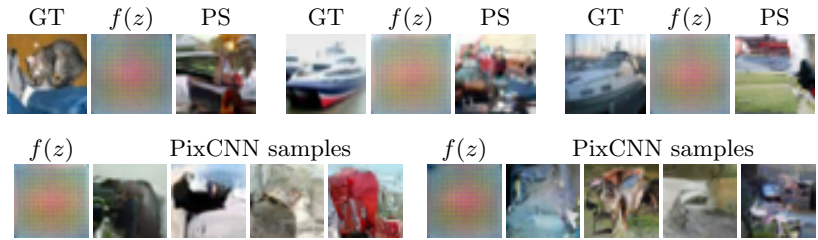
\begin{figure}[t]
		\begin{center}
			\input{./figures/test_rebuttal_control_rec_small.tex}
		\end{center}
	       \caption{Auxiliary reconstructions obtained after dropping the auxilliary loss. (GT) denotes ground truth images unseen during training, $f(z)$ is the corresponding intermediate reconstruction, (PS) denotes pixelCNN samples, conditionned on $f(z)$.}
	       \label{fig:control_reconstructions}
	\end{figure}
{}

\ble{The fact that the autoregressive decoder ignores the latent variables could be attributed to optimization challenges, as explained in \ref{sec:bits_back}. In that case, the auxilliary loss could be used as an initialization scheme only, to guide the model towards a good use of the latent variables. To evaluate this we perform a control experiment where during training we first optimize our objective function in Eq.\ (\ref{eq:agave}), \ie including the auxiliary reconstruction term, and then switch to optimize the standard objective function of  Eq.\ (\ref{eq:elboAR}) without the auxiliary term. We proceed by training the full model to convergence then removing the auxiliary loss and fine-tuning from there. 
\fig{control_reconstructions} displays ground-truth images, with corresponding auxiliary reconstructions and conditional samples, as well as pure samples. The reconstructions have become meaningless and independent from the ground truth images. The samples display the same behavior: for each auxiliary representation four samples from the autoregressive component are displayed and they are independent from one another.
Quantitatively, the KL cost immediately drops to zero when removing the auxiliary loss, in approximately two thousand steps of gradient descent.
The approximate posterior immediately collapses to the prior and the pixel CNN samples become independent of the latent variables. 
This is the behavior predicted by the analysis of \citet{chen17iclr}: the autoregressive decoder is sufficiently expressive that it suffers from using the latent variables.}
\end{subsection}

\subsection{Effect of different auxiliary images.}

	\begin{figure}[]
		\begin{center}
			\begin{minipage}[b]{0.48\textwidth}
				\begin{center}
					\scalebox{1.1}{
					\input{./figures/fig5a.tex}}\\
					(a)
				\end{center}
			\end{minipage}			\hfill
			\begin{minipage}[b]{0.45\textwidth}
				\centering
				\includegraphics[width=\linewidth]{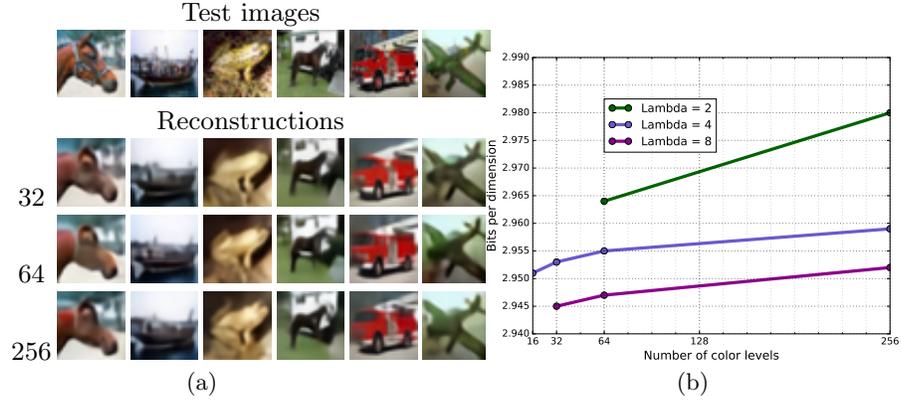}\\
				(b)
			\end{minipage}
		\end{center}
		\caption{
			Impact of the color quantization in the auxiliary image. 
			(a) Reconstructions of the VAE decoder for different quantization levels ($\lambda=8$).
			(b) BPD as a function of the quantization level.
		}
		\label{fig:bin_plots}
	\end{figure}
{}

	\begin{figure}[]
		\begin{center}
			\begin{minipage}[b]{0.48\textwidth}
				\begin{center}
																				\input{./figures/grey_images_latex_tile_grey.tex}\\
					(a)
				\end{center}
			\end{minipage}
			\hfill
			\begin{minipage}[b]{0.48\textwidth}
				\begin{center}
																				\input{./figures/bin_samples_pix0_small.tex}\\
					(b)
				\end{center}
			\end{minipage}
		\end{center}
		\begin{center}
			\begin{minipage}[b]{0.48\textwidth}
				\begin{center}
					\input{./figures/coarse_16_new_small.tex}\\
																				(c)
				\end{center}
			\end{minipage}
			\hfill
			\begin{minipage}[b]{0.48\textwidth}
				\begin{center}
										\input{./figures/pixout_new8x_pixVAE.tex}\\
					(d)
				\end{center}
			\end{minipage}
		\end{center}
				\caption{
			Samples from models trained with grayscale auxiliary images with 16 color levels (a), 32$\times$32 auxiliary images with 32 color levels (b), and at reduced resolutions of 16$\times$16 (c) and 8$\times$8 pixels (d) with 256 color levels.
			For each model the auxilliary representation $f(\bz)$, with $\bz$ sampled from the prior, is displayed above the corresponding conditional pixelCNN sample.
		}
		\label{fig:bin_samples}
	\end{figure}
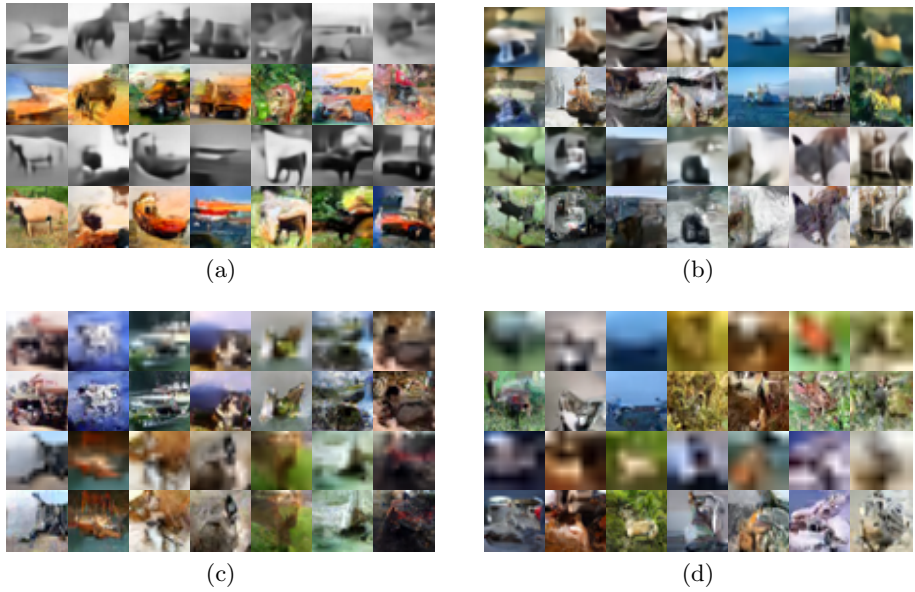
{}

\ble{We  assess the effect of using coarser RGB  quantizations, lower spatial resolutions, and grayscale in the auxiliary image. 
All three make the VAE reconstruction task easier, and transfer the task of modeling color nuances and/or spatial detail to the pixelCNN.}

The VAE reconstructions in \fig{bin_plots} (a) obtained using coarser color quantization carry less detail than reconstructions based on the original images using 256 color values,  as expected.
To understand the relatively small impact of the quantization level on the reconstruction, recall that the VAE decoder outputs the continuous means of the logistic distributions regardless of the quantization level. Only the reconstruction loss is impacted by the quantization level via the computation of the probabilities over the discrete color levels in \Eq{sigmoid}.
In \fig{bin_plots} (b) we observe small but consistent gains in the BPD metric as the number of color bins is reduced, showing that it is more effective to model color nuances using the pixelCNN, rather than the latent variables.
We trained models with auxiliary images down-sampled to  16$\times$16 and 8$\times$8 pixels, which yield  2.94 and 2.93 BPD, respectively.
This is comparable to the 2.92 BPD obtained using our best model at scale 32$\times$32. \red{We also trained models with 4-bit per pixel grayscale auxiliary images, as in \citet{kolesnikov17icml}. While the grayscale auxilliary images are subjectively the ones that have the best global structure, the results are still qualitatively inferior to those obtained by \citet{kolesnikov17icml} with a pixelCNN modelling grayscale images. Our model does, however, achieve better quantitative performance at 2.93 BPD.}
In \fig{bin_samples} (a) we show samples obtained using models trained with 4-bit per pixel grayscale auxiliary images,
\ble{in \fig{bin_samples} (b) with 32 color levels in the auxiliary image}, and in  \fig{bin_samples} (c) and (d) with auxiliary images of size 16$\times$16 and 8$\times$8.
The samples are qualitatively comparable, showing that in all cases the pixelCNN is able to compensate the less detailed outputs of the VAE decoder \ble{and that our framework can be used with a variety of intermediate reconstruction losses.}

\begin{section}{Conclusion}
We presented a new approach to training generative image models that combine a latent variable structure with an autoregressive model component. 
Unlike prior approaches, it  does not require careful architecture design to
trade-off how much is modeled by latent variables and the autoregressive decoder.
Instead, this trade-off can be controlled using a regularization parameter 
and choice of auxiliary target images. 
We obtain quantitative performance on par with the state of the art on CIFAR10, 
and samples from our model exhibit globally coherent structure as well as fine details. 
\end{section}

\subsubsection*{Acknowledgments}
\medskip
{
This work has been partially supported by the grant ANR-16-CE23-0006 
``Deep in
France'' and LabEx PERSYVAL-Lab (ANR-11-LABX-0025-01).
}

\bibliography{biblio,jjv}
\bibliographystyle{splncs03}
\end{document}

%% file: figures/dec_mismatch0.tex
\setlength{\tabcolsep}{0pt}
\begin{tabular}{cccccccc} 
VR & \includegraphics{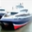} &
\includegraphics{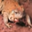} &
\includegraphics{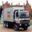} &
\includegraphics{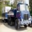} &
\includegraphics{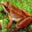} &
\includegraphics{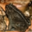}\\
PS & \includegraphics{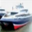} &
\includegraphics{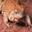} &
\includegraphics{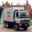} &
\includegraphics{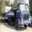} &
\includegraphics{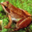} &
\includegraphics{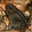}\\
\end{tabular} 

%% file: figures/dec_mismatch1.tex
\setlength{\tabcolsep}{0pt}
\begin{tabular}{cccccccc} 
VR & \includegraphics{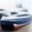} &
\includegraphics{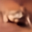} &
\includegraphics{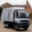} &
\includegraphics{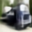} &
\includegraphics{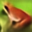} &
\includegraphics{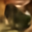}\\
PS & \includegraphics{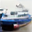} &
\includegraphics{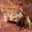} &
\includegraphics{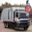} &
\includegraphics{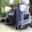} &
\includegraphics{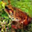} &
\includegraphics{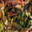}\\
\end{tabular} 

%% file: figures/dec_mismatch2.tex
\setlength{\tabcolsep}{0pt}
\begin{tabular}{cccccccc} 
VR & \includegraphics{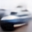} &
\includegraphics{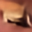} &
\includegraphics{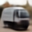} &
\includegraphics{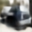} &
\includegraphics{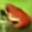} &
\includegraphics{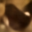}\\
PS & \includegraphics{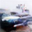} &
\includegraphics{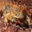} &
\includegraphics{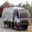} &
\includegraphics{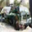} &
\includegraphics{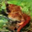} &
\includegraphics{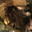}\\
\end{tabular} 

%% file: figures/resp_cond_v20.tex
\setlength{\tabcolsep}{0pt}
\begin{tabular}{cccccccc} 
\includegraphics{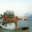} &
\includegraphics{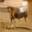} &
\includegraphics{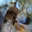} &
\includegraphics{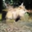} &
\includegraphics{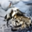} &
\includegraphics{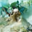} &
VS\\\includegraphics{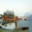} &
\includegraphics{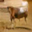} &
\includegraphics{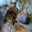} &
\includegraphics{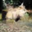} &
\includegraphics{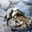} &
\includegraphics{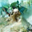} &
PS\\\end{tabular} 

%% file: figures/resp_cond_v21.tex
\setlength{\tabcolsep}{0pt}
\begin{tabular}{cccccccc} 
\includegraphics{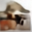} &
\includegraphics{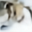} &
\includegraphics{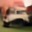} &
\includegraphics{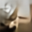} &
\includegraphics{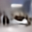} &
\includegraphics{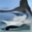} &
VS\\\includegraphics{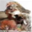} &
\includegraphics{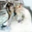} &
\includegraphics{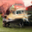} &
\includegraphics{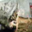} &
\includegraphics{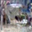} &
\includegraphics{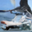} &
PS\\\end{tabular} 

%% file: figures/resp_cond_v22.tex
\setlength{\tabcolsep}{0pt}
\begin{tabular}{cccccccc} 
\includegraphics{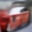} &
\includegraphics{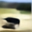} &
\includegraphics{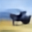} &
\includegraphics{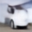} &
\includegraphics{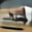} &
\includegraphics{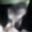} &
VS\\
\includegraphics{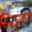} &
\includegraphics{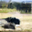} &
\includegraphics{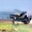} &
\includegraphics{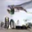} &
\includegraphics{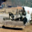} &
\includegraphics{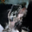} &
PS\\
\end{tabular} 

%% file: figures/test_rebuttal_control_rec_small.tex
\setlength{\tabcolsep}{0pt}
\begin{tabular}{c@{\hskip0.5ex}c@{\hskip0.5ex}c@{\hskip3.0ex}c@{\hskip0.5ex}c@{\hskip0.5ex}c@{\hskip3.0ex}c@{\hskip0.5ex}c@{\hskip0.5ex}c@{\hskip0.5ex}} 
	GT & $f(z)$ & PS & GT & $f(z)$ & PS & GT & $f(z)$ & PS\\ 
\includegraphics[scale=1.3]{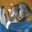} &
\includegraphics[scale=1.3]{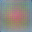} &
\includegraphics[scale=1.3]{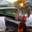} &
\includegraphics[scale=1.3]{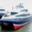} &
\includegraphics[scale=1.3]{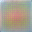} &
\includegraphics[scale=1.3]{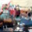} &
\includegraphics[scale=1.3]{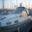} &
\includegraphics[scale=1.3]{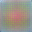} &
\includegraphics[scale=1.3]{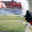}\\[1ex]
\end{tabular} 
\setlength{\tabcolsep}{0pt}

\begin{tabular}{c@{\hskip0.5ex}c@{\hskip0.5ex}c@{\hskip0.5ex}c@{\hskip0.5ex}c@{\hskip2.5ex}c@{\hskip0.5ex}c@{\hskip0.5ex}c@{\hskip0.5ex}c@{\hskip0.5ex}c@{\hskip0.5ex}} 
	$f(z)$ & \multicolumn{4}{c}{PixCNN samples} & $f(z)$ & \multicolumn{4}{c}{PixCNN samples} \\
\includegraphics[scale=1.2]{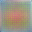} &
\includegraphics[scale=1.2]{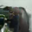} &
\includegraphics[scale=1.2]{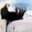} &
\includegraphics[scale=1.2]{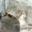} &
\includegraphics[scale=1.2]{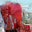} &
\includegraphics[scale=1.2]{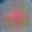} &
\includegraphics[scale=1.2]{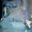} &
\includegraphics[scale=1.2]{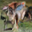} &
\includegraphics[scale=1.2]{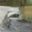} &
\includegraphics[scale=1.2]{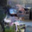}\\[3ex]
\end{tabular} 

%% file: figures/fig5a.tex
\setlength{\tabcolsep}{1pt}
{
\footnotesize
\begin{tabular}{ccccccc} 
\multicolumn{7}{c}{Test images}\\
 & 
\includegraphics{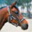} &
\includegraphics{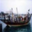} &
\includegraphics{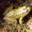} &
\includegraphics{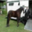} &
\includegraphics{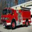} &
\includegraphics{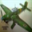}\\
\multicolumn{7}{c}{Reconstructions}
\\
32 &
\includegraphics{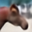} &
\includegraphics{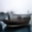} &
\includegraphics{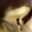} &
\includegraphics{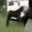} &
\includegraphics{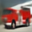} &
\includegraphics{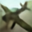}\\
64 & 
\includegraphics{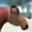} &
\includegraphics{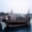} &
\includegraphics{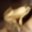} &
\includegraphics{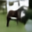} &
\includegraphics{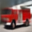} &
\includegraphics{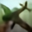}\\
256 & 
\includegraphics{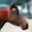} &
\includegraphics{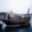} &
\includegraphics{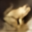} &
\includegraphics{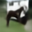} &
\includegraphics{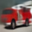} &
\includegraphics{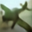}
\end{tabular} 
}

%% file: figures/grey_images_latex_tile_grey.tex
\setlength{\tabcolsep}{0pt}
\begin{tabular}{ccccccc} 
\includegraphics{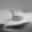} &
\includegraphics{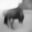} &
\includegraphics{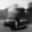} &
\includegraphics{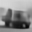} &
\includegraphics{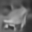} &
\includegraphics{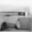} &
\includegraphics{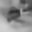}\\[-0.85ex]
\includegraphics{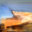} &
\includegraphics{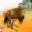} &
\includegraphics{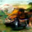} &
\includegraphics{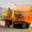} &
\includegraphics{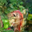} &
\includegraphics{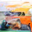} &
\includegraphics{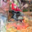}\\[-0.85ex]
\includegraphics{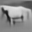} &
\includegraphics{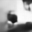} &
\includegraphics{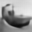} &
\includegraphics{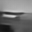} &
\includegraphics{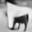} &
\includegraphics{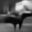} &
\includegraphics{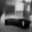}\\[-0.85ex]
\includegraphics{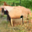} &
\includegraphics{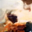} &
\includegraphics{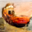} &
\includegraphics{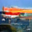} &
\includegraphics{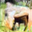} &
\includegraphics{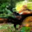} &
\includegraphics{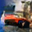}\\
\end{tabular} 

%% file: figures/bin_samples_pix0_small.tex
\setlength{\tabcolsep}{0pt}
\begin{tabular}{cccccccc} 
\includegraphics{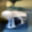} &
\includegraphics{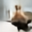} &
\includegraphics{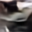} &
\includegraphics{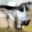} &
\includegraphics{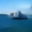} &
\includegraphics{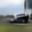} &
\includegraphics{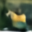}\\[-1ex]
\includegraphics{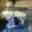} &
\includegraphics{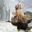} &
\includegraphics{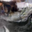} &
\includegraphics{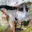} &
\includegraphics{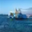} &
\includegraphics{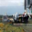} &
\includegraphics{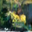}\\[-1ex]
\includegraphics{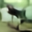} &
\includegraphics{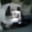} &
\includegraphics{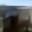} &
\includegraphics{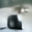} &
\includegraphics{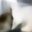} &
\includegraphics{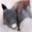} &
\includegraphics{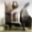}\\[-1ex]
\includegraphics{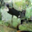} &
\includegraphics{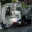} &
\includegraphics{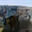} &
\includegraphics{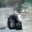} &
\includegraphics{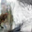} &
\includegraphics{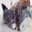} &
\includegraphics{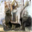}\\
\end{tabular} 

%% file: figures/coarse_16_new_small.tex
\setlength{\tabcolsep}{0pt}
\begin{tabular}{cccccccc} 
\includegraphics{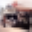} &
\includegraphics{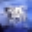} &
\includegraphics{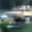} &
\includegraphics{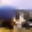} &
\includegraphics{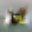} &
\includegraphics{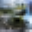} &
\includegraphics{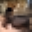}\\[-1ex]
\includegraphics{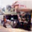} &
\includegraphics{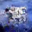} &
\includegraphics{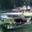} &
\includegraphics{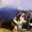} &
\includegraphics{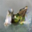} &
\includegraphics{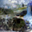} &
\includegraphics{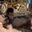}\\[-1ex]
\includegraphics{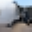} &
\includegraphics{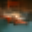} &
\includegraphics{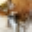} &
\includegraphics{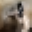} &
\includegraphics{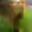} &
\includegraphics{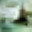} &
\includegraphics{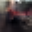}\\[-1ex]
\includegraphics{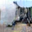} &
\includegraphics{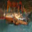} &
\includegraphics{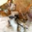} &
\includegraphics{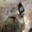} &
\includegraphics{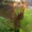} &
\includegraphics{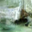} &
\includegraphics{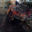}\\
\end{tabular} 

%% file: figures/pixout_new8x_pixVAE.tex
\setlength{\tabcolsep}{0pt}
\begin{tabular}{cccccccc} 
\includegraphics{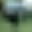} &
\includegraphics{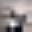} &
\includegraphics{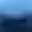} &
\includegraphics{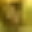} &
\includegraphics{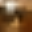} &
\includegraphics{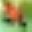} &
\includegraphics{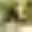}\\[-1ex]
\includegraphics{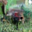} &
\includegraphics{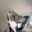} &
\includegraphics{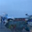} &
\includegraphics{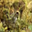} &
\includegraphics{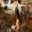} &
\includegraphics{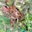} &
\includegraphics{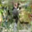}\\[-1ex]
\includegraphics{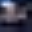} &
\includegraphics{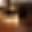} &
\includegraphics{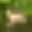} &
\includegraphics{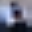} &
\includegraphics{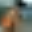} &
\includegraphics{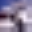} &
\includegraphics{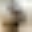}\\[-1ex]
\includegraphics{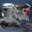} &
\includegraphics{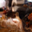} &
\includegraphics{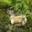} &
\includegraphics{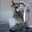} &
\includegraphics{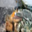} &
\includegraphics{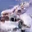} &
\includegraphics{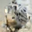}\\
\end{tabular} 

%% file: figures/test_rebuttal_array_small.tex
\setlength{\tabcolsep}{0pt}
\begin{tabular}{c@{\hskip2.5ex}c@{\hskip0.5ex}c@{\hskip0.5ex}c@{\hskip0.5ex}c@{\hskip0.5ex}c@{\hskip0.5ex}c@{\hskip0.5ex}c@{\hskip0.5ex}c@{\hskip0.5ex}c@{\hskip0.5ex}} 
	$f(\bz)$ & \multicolumn{9}{c}{Conditional PixelCNN samples} \\
\includegraphics[scale=1.4]{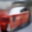} &
\includegraphics[scale=1.4]{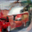} &
\includegraphics[scale=1.4]{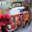} &
\includegraphics[scale=1.4]{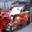} &
\includegraphics[scale=1.4]{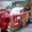} &
\includegraphics[scale=1.4]{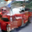} &
\includegraphics[scale=1.4]{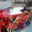} &
\includegraphics[scale=1.4]{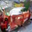} &
\includegraphics[scale=1.4]{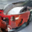} &
\includegraphics[scale=1.4]{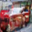}\\[2ex]
\includegraphics[scale=1.4]{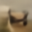} &
\includegraphics[scale=1.4]{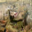} &
\includegraphics[scale=1.4]{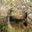} &
\includegraphics[scale=1.4]{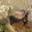} &
\includegraphics[scale=1.4]{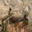} &
\includegraphics[scale=1.4]{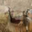} &
\includegraphics[scale=1.4]{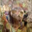} &
\includegraphics[scale=1.4]{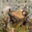} &
\includegraphics[scale=1.4]{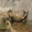} &
\includegraphics[scale=1.4]{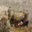}\\[2ex]
\includegraphics[scale=1.4]{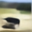} &
\includegraphics[scale=1.4]{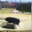} &
\includegraphics[scale=1.4]{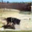} &
\includegraphics[scale=1.4]{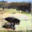} &
\includegraphics[scale=1.4]{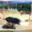} &
\includegraphics[scale=1.4]{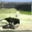} &
\includegraphics[scale=1.4]{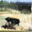} &
\includegraphics[scale=1.4]{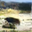} &
\includegraphics[scale=1.4]{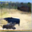} &
\includegraphics[scale=1.4]{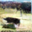}\\[2ex]
\includegraphics[scale=1.4]{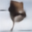} &
\includegraphics[scale=1.4]{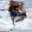} &
\includegraphics[scale=1.4]{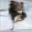} &
\includegraphics[scale=1.4]{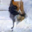} &
\includegraphics[scale=1.4]{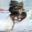} &
\includegraphics[scale=1.4]{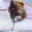} &
\includegraphics[scale=1.4]{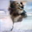} &
\includegraphics[scale=1.4]{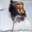} &
\includegraphics[scale=1.4]{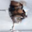} &
\includegraphics[scale=1.4]{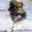}\\[2ex]
\end{tabular}

%
%

%% file: figures/vae_kl1_samplesvae_kl1.tex
\setlength{\tabcolsep}{0pt}
\begin{tabular}{cccc} 
\includegraphics{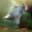} &
\includegraphics{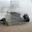} &
\includegraphics{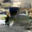} &
\includegraphics{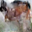}\\[-1ex]
\includegraphics{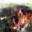} &
\includegraphics{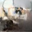} &
\includegraphics{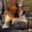} &
\includegraphics{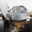}\\[-1ex]
\includegraphics{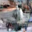} &
\includegraphics{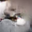} &
\includegraphics{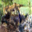} &
\includegraphics{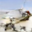}\\[-1ex]
\includegraphics{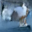} &
\includegraphics{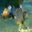} &
\includegraphics{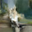} &
\includegraphics{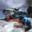}\\[-1ex]
\includegraphics{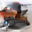} &
\includegraphics{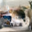} &
\includegraphics{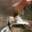} &
\includegraphics{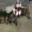}\\[-1ex]
\includegraphics{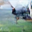} &
\includegraphics{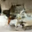} &
\includegraphics{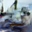} &
\includegraphics{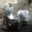}\\[-1ex]
\includegraphics{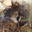} &
\includegraphics{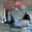} &
\includegraphics{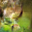} &
\includegraphics{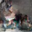}\\[-1ex]
\includegraphics{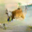} &
\includegraphics{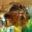} &
\includegraphics{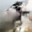} &
\includegraphics{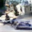}\\
\end{tabular} 

%% file: figures/pixcnnpp_samplespixcnn_pp.tex
\setlength{\tabcolsep}{0pt}
\begin{tabular}{cccc} 
\includegraphics{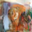} &
\includegraphics{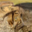} &
\includegraphics{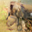} &
\includegraphics{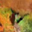}\\[-1ex]
\includegraphics{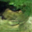} &
\includegraphics{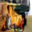} &
\includegraphics{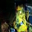} &
\includegraphics{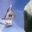}\\[-1ex]
\includegraphics{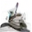} &
\includegraphics{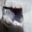} &
\includegraphics{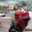} &
\includegraphics{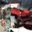}\\[-1ex]
\includegraphics{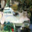} &
\includegraphics{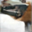} &
\includegraphics{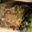} &
\includegraphics{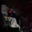}\\[-1ex]
\includegraphics{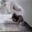} &
\includegraphics{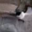} &
\includegraphics{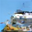} &
\includegraphics{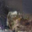}\\[-1ex]
\includegraphics{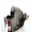} &
\includegraphics{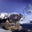} &
\includegraphics{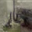} &
\includegraphics{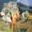}\\[-1ex]
\includegraphics{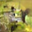} &
\includegraphics{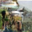} &
\includegraphics{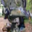} &
\includegraphics{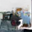}\\[-1ex]
\includegraphics{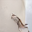} &
\includegraphics{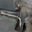} &
\includegraphics{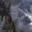} &
\includegraphics{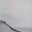}\\
\end{tabular} 

%% file: figures/pix_out_samples_figure_kl8_bin32_small.tex
\setlength{\tabcolsep}{0pt}
\begin{tabular}{cccccccc} 
\includegraphics{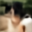} &
\includegraphics{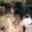} &
\includegraphics{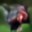} &
\includegraphics{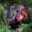} &
\includegraphics{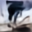} &
\includegraphics{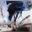} \\[-1ex]
\includegraphics{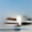} &
\includegraphics{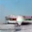} &
\includegraphics{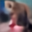} &
\includegraphics{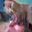} &
\includegraphics{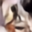} &
\includegraphics{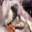} \\[-1ex]
\includegraphics{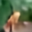} &
\includegraphics{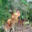} &
\includegraphics{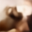} &
\includegraphics{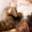} &
\includegraphics{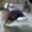} &
\includegraphics{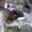} \\[-1ex]
\includegraphics{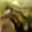} &
\includegraphics{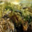} &
\includegraphics{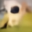} &
\includegraphics{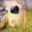} &
\includegraphics{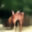} &
\includegraphics{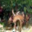} \\[-1ex]
\includegraphics{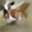} &
\includegraphics{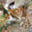} &
\includegraphics{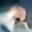} &
\includegraphics{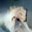} &
\includegraphics{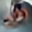} &
\includegraphics{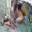} \\[-1ex]
\includegraphics{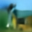} &
\includegraphics{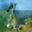} &
\includegraphics{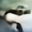} &
\includegraphics{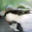} &
\includegraphics{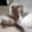} &
\includegraphics{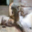} \\[-1ex]
\includegraphics{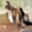} &
\includegraphics{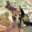} &
\includegraphics{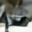} &
\includegraphics{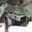} &
\includegraphics{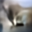} &
\includegraphics{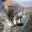} \\[-1ex]
\includegraphics{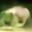} &
\includegraphics{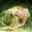} &
\includegraphics{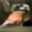} &
\includegraphics{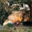} &
\includegraphics{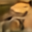} &
\includegraphics{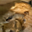} \\
\end{tabular} 

%% file: figures/pixVAE_images_latent_interpolation_latex_tile_both.tex
\setlength{\tabcolsep}{0pt}
\begin{tabular}{ccccccccccccccc} 
\includegraphics{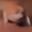} &
\includegraphics{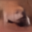} &
\includegraphics{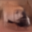} &
\includegraphics{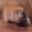} &
\includegraphics{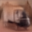} &
\includegraphics{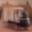} &
\includegraphics{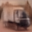} &
\includegraphics{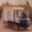} &
\includegraphics{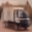} &
\includegraphics{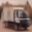} &
\includegraphics{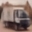} &
\includegraphics{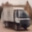} &
\includegraphics{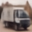} &
\includegraphics{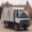} &
\includegraphics{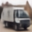}\\[-0.85ex]
\includegraphics{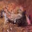} &
\includegraphics{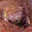} &
\includegraphics{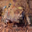} &
\includegraphics{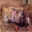} &
\includegraphics{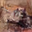} &
\includegraphics{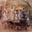} &
\includegraphics{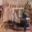} &
\includegraphics{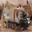} &
\includegraphics{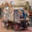} &
\includegraphics{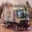} &
\includegraphics{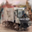} &
\includegraphics{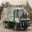} &
\includegraphics{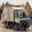} &
\includegraphics{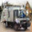} &
\includegraphics{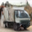}\\[-0.85ex]
\includegraphics{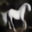} &
\includegraphics{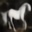} &
\includegraphics{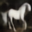} &
\includegraphics{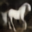} &
\includegraphics{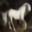} &
\includegraphics{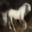} &
\includegraphics{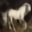} &
\includegraphics{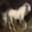} &
\includegraphics{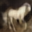} &
\includegraphics{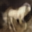} &
\includegraphics{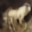} &
\includegraphics{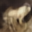} &
\includegraphics{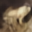} &
\includegraphics{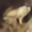} &
\includegraphics{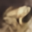}\\[-0.85ex]
\includegraphics{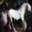} &
\includegraphics{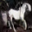} &
\includegraphics{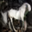} &
\includegraphics{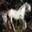} &
\includegraphics{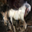} &
\includegraphics{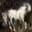} &
\includegraphics{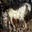} &
\includegraphics{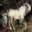} &
\includegraphics{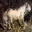} &
\includegraphics{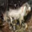} &
\includegraphics{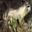} &
\includegraphics{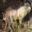} &
\includegraphics{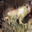} &
\includegraphics{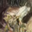} &
\includegraphics{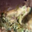}\\[-0.85ex]
\includegraphics{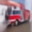} &
\includegraphics{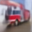} &
\includegraphics{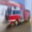} &
\includegraphics{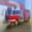} &
\includegraphics{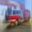} &
\includegraphics{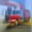} &
\includegraphics{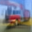} &
\includegraphics{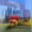} &
\includegraphics{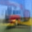} &
\includegraphics{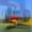} &
\includegraphics{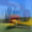} &
\includegraphics{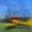} &
\includegraphics{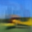} &
\includegraphics{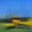} &
\includegraphics{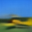}\\
\includegraphics{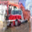} &
\includegraphics{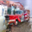} &
\includegraphics{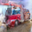} &
\includegraphics{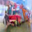} &
\includegraphics{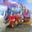} &
\includegraphics{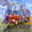} &
\includegraphics{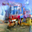} &
\includegraphics{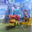} &
\includegraphics{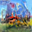} &
\includegraphics{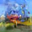} &
\includegraphics{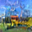} &
\includegraphics{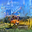} &
\includegraphics{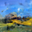} &
\includegraphics{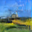} &
\includegraphics{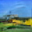}\\
\end{tabular} 

%% file: figures/pixvae_images_grey_images_latex_tile_grey.tex
\setlength{\tabcolsep}{0pt}
\begin{tabular}{ccccccc} 
\includegraphics{./images/grey_images_individual_images/img46.png} &
\includegraphics{./images/grey_images_individual_images/img64.png} &
\includegraphics{./images/grey_images_individual_images/img120.png} &
\includegraphics{./images/grey_images_individual_images/img148.png} &
\includegraphics{./images/grey_images_individual_images/img182.png} &
\includegraphics{./images/grey_images_individual_images/img304.png} &
\includegraphics{./images/grey_images_individual_images/img368.png}\\[-0.85ex]
\includegraphics{./images/grey_images_individual_images/img47.png} &
\includegraphics{./images/grey_images_individual_images/img65.png} &
\includegraphics{./images/grey_images_individual_images/img121.png} &
\includegraphics{./images/grey_images_individual_images/img149.png} &
\includegraphics{./images/grey_images_individual_images/img183.png} &
\includegraphics{./images/grey_images_individual_images/img305.png} &
\includegraphics{./images/grey_images_individual_images/img369.png}\\[-0.85ex]
\includegraphics{./images/grey_images_individual_images/img54.png} &
\includegraphics{./images/grey_images_individual_images/img90.png} &
\includegraphics{./images/grey_images_individual_images/img144.png} &
\includegraphics{./images/grey_images_individual_images/img180.png} &
\includegraphics{./images/grey_images_individual_images/img202.png} &
\includegraphics{./images/grey_images_individual_images/img310.png} &
\includegraphics{./images/grey_images_individual_images/img398.png}\\[-0.85ex]
\includegraphics{./images/grey_images_individual_images/img55.png} &
\includegraphics{./images/grey_images_individual_images/img91.png} &
\includegraphics{./images/grey_images_individual_images/img145.png} &
\includegraphics{./images/grey_images_individual_images/img181.png} &
\includegraphics{./images/grey_images_individual_images/img203.png} &
\includegraphics{./images/grey_images_individual_images/img311.png} &
\includegraphics{./images/grey_images_individual_images/img399.png}\\
\end{tabular} 